# It is time for Factor Graph Optimization for GNSS/INS Integration: Comparison between FGO and EKF

Weisong Wen, Tim Pfeifer, Xiwei Bai, Li-Ta Hsu*

*Abstract*—**The recently proposed factor graph optimization (FGO) is adopted to integrate GNSS/INS attracted lots of attention and improved the performance over the existing EKF-based GNSS/INS integrations. However, a comprehensive comparison of those two GNSS/INS integration schemes in the urban canyon is not available. Moreover, the performance of the FGO-based GNSS/INS integration rely heavily on the size of the window of optimization. Effectively tuning the window size is still an open question. To fill this gap, this paper evaluates both loosely and tightly-coupled integrations using both EKF and FGO via the challenging dataset collected in the urban canyon. The detailed analysis of the results for the advantages of the FGO is also given in this paper by degenerating the FGO-based estimator to an "EKF like estimator". More importantly, we analyze the effects of window size against the performance of FGO, by considering both the GNSS pseudorange error distribution and environmental conditions.**

## I. INTRODUCTION

GNSS can provide all-weather and globally referenced positioning in outdoor environments. Sufficient positioning performance can be obtained in scenes with high sky visibility. However, the accuracy of GNSS positioning can be severely degraded in urban canyons with tall buildings, due to the multipath effects and Non-line-of-sight (NLOS) [1] receptions caused by reflections and blockage from buildings. The inertial navigation system (INS) [2, 3] can provide relative linear acceleration and angular velocity measurements at a high output frequency. It is also less dependent on environmental conditions. However, the INS suffers from error accumulation over time. Therefore, the GNSS/INS integration is a promising solution for vehicular positioning that makes use of their complementariness.

GNSS/INS integration frameworks are reviewed in [4]. The major integration solutions are loosely-coupled (LC) [5-7], tightly-coupled (TC) [8], and ultra-tightly-coupled (UTC) [9] integration. The UTC integration requires the change of the baseband signal processing of the GNSS receiver, which is usually not accessible for GNSS/INS integrated system developers. The major difference between the LC and TC integrations is the domain of measurements used in the integration. In the LC GNSS/INS integration, the position and velocity estimated by the GNSS receiver are directly incorporated with the INS navigation solution. In contrast, raw GNSS measurements, such as pseudorange, Doppler frequency, and carrier-phase are used for a TC integration. The TC GNSS/INS integration can obtain better performance than the LC one based on an extended Kalman filter (EKF), which is shown in [8]. The main reason behind this is that the quality of GNSS measurements can be modeled better [8]. In short, the TC GNSS/INS integration using EKF is a popular solution for the existing applications.

The recently proposed factor graph optimization (FGO) [10] is a popular approach in the robotics field. It is usually used in the development of the visual- [11] or LiDAR simultaneous localization and mapping (SLAM) [12] to integrate diverse sensor measurements via non-linear optimization. It has huge potential to also be used for GNSS/INS integration which attracts lots of attention [13-16] recently. However, an integrated performance comparison of the recently proposed FGO and EKF for the low-cost GNSS and INS integration using the challenging dataset is not available in the existing work. Moreover, according to the findings in [15, 17], the accuracy and efficiency of the FGO-based GNSS/INS integration rely heavily on the size of the window of optimization. Effectively adapting the window size is still an open question. The objective of this paper is therefore to compare the performance of the EKF and the FGO both in LC and TC of GNSS/INS integrations. Importantly, we analyze the effects of window size against the performance of FGO based on the validated dataset. Although the FGO shows better performance in the SLAM research field, it is still not popular in the navigation field due to the increased time efficiency of FGO. Therefore, we also show the time efficiency of EKF and FGO based on the tested dataset. To the best of the authors' knowledge, this is the first paper that compares the performance of the EKF and the FGO both in LC and TC of GNSS/INS integrations using a real dataset collected in deep urban canyons. The contributions of this paper are as follows:

(1) The performances of the four combinations of GNSS/INS are compared using the real dataset collected in the urban canyon of Hong Kong. The detailed analysis of the

Weisong Wen, Xiwei Bai and L.T. Hsu, are with Hong Kong Polytechnic University, Hong Kong (correspondence e-mail: lt.hsu@polyu.edu.hk).

Tim Pfeifer is with Chemnitz University of Technology, Germany (e-mail: tim.pfeifer@etit.tu-chemnitz.de).

results for the advantages of the FGO is also given in this paper based on real data evaluation.

(2) The effects of the window size against the performance of FGO, based on the validated dataset are analyzed by considering both the GNSS pseudorange measurement residuals distribution and environmental conditions

The remainder of this paper is structured as follows. Firstly, the related work of EKF and FGO for GNSS/INS integration is reviewed in Section II. The methodologies of the evaluations for the four GNSS/INS integration are given in Section III. The experimental evaluations of the four integrations are presented in Section IV. Finally, the conclusions and future work are given in Section V.

## II. Related Works

### A. Extended Kalman Filter (EKF) in GNSS/INS Integration

The Bayesian filter [18] has dominated the GNSS/INS integration since the early millennium. Kalman filter (KF) [19], EKF [20, 21], and unscented Kalman filter (UKF) [22] are utterly popular, due to their maturity and computational efficiency in implementations. Numerous researches [23-25] have been conducted to integrated GNSS/INS sensors using the popular EKF estimator. Numerous real application shows the robustness and effectiveness of EKF in GNSS when the measurement quality is decent and its error noise is properly modeled. For example, the EKF-based GNSS/INS integration can work well in a sparse or open area with decent sky visibility. Unfortunately, the performance of GNSS/INS integration via EKF is significantly degraded in urban canyons due to the tall buildings, which cause numerous GNSS outlier measurements [26, 27].

According to the finding in [17], the GNSS measurement at the current epoch and historical is highly time-correlated in the urban canyons. However, conventional EKF is assumed to follow the first-order Markov chain [28], which fails to fully make use of historical information. It heavily relies on the state at the previous epoch and measurements at the current epoch [28, 29]. Therefore, EKF-based sensor fusion fails to take advantage of previous information. From a mathematical perspective, the EKF only evaluates the Jacobians at a single time step (single iteration) due to the Markov assumption to achieve its recursive form. It does not maintain enough preceding measurements (redundant information) to resist outliers [30]. When an outliner is miss-judged as a healthy measurement and respective uncertainty is not properly modeled, EKF is very likely to be miss-leaded, which is unacceptable for applications requiring accurate positioning services, such as unmanned aerial vehicles (UAV) [31] and autonomous driving vehicles (ADV) [32]. Whereas, the outlier is not an occasional case for GNSS positioning in urban canyons. To fully make use of the historical information in the EKF estimator, one possible solution is to augment the historical information into the state vector. However, this will significantly increase the size of the estimator. Moreover, the convergence of the EKF estimator will be significantly decreased [33]. Recently, the multi-state constrained Kalman filter (MSCKF) [34] was proposed in the visual SLAM field to integrated the information from INS and camera. The MSCKF updates the states based on the geometry constraints of feature measurements inside the sliding window (involving history information). However, the states of the features are eliminated from the MSCKF via the nullspace matrix to reduce the size of the states. In other words, the MSCKF did not fully make use of historical information.

Besides, the performance of the EKF [20] relies on the accurate linearization due to the non-linearity of the observation function, and only single linearization is performed in the EKF. As a result, the accuracy of linearization relies heavily on the initial guess of the state. To solve this problem, the iterated Kalman filter (IKF) [35] is proposed to perform multiple iterations during the update step using the Gauss-Newton method. The IKF can effectively help to mitigate the error from the linearization steps. Whereas, the historical information is still not well utilized.

### B. Factor Graph Optimization (FGO) in GNSS/INS Integration

The recently proposed FGO [36] formulation opens a new window for multi-sensor integration [37-39]. It is represented by a probabilistic graphical model with various nodes associated with system states, and factors representing the measurements. The factor graph encodes the posterior probability of the states over time. The factor graph considers both the historical measurements and system updates to optimize the full state set, which is different from the conventional EKF-based integration. In this case, the historical information is employed in FGO. Besides, after encoding all the measurements and states into a factor graph, the sensor fusion problem is solved iteratively via optimization using the Gauss-Newton method. Therefore, the error arising from the linearization steps can be mitigated accordingly. Moreover, the FGO is capable of coping with the delayed measurements as the delayed measurements are simply additional sources of factors that get added to the graph. The factor graph is updated when new measurements occur. As a result, a recent fashion is to make use of the FGO in kinds of GNSS challenges scenes [40-43]. Interestingly, the recent work in [44] shows the strong potential of FGO in sensor fusion even when the sensor noise is modeled with the non-Gaussian distribution.

One of the applications of the FGO, which recently attracts lots of attention in the navigation field, is the integration of GNSS and INS [13-16]. The team [45] from the Georgia Institute of Technology shows the advantage of FGO in LC GNSS/INS integration compared with the EKF estimator using simulated data. However, only limited improvement is obtained in the loosely-coupled integration. The team [13] from Tsinghua university goes a further step where the TC GNSS/INS integration is studied and significantly improved performance is obtained compared with the EKF estimator. Unfortunately, both of the teams share the same drawback of only evaluating the simulated data and the magnitude of GNSS measurement error is limited. Inspired by the novel work of both teams, we go step further where we study the performance of TC GNSS/INS/Fisheye camera integration [14] using FGO based on the challenging dataset collected in the deep urban canyons of Hong Kong. The fisheye camera is innovatively employed to model the uncertainty of GNSS pseudorange measurements before its integration with INS using FGO. The results show the outperforming performance of TC FGO compared with the TC EKF estimator. However, the performance evaluations of LC integrations for both EKF and FGO are not available in [14].

In parallel, interestingly, the team [15] from the University of California, Riverside (UCR) proposed an optimization-based sliding window for differential GNSS (DGNSS) and INS integration, the Contemplative Real-Time (CRT) method. A Gauss-Newton method is then applied to optimize the states inside the window based on the given sensor measurements. The CRT shares the same theoretical basis with the factor graph. The first work in [16] presents the CRT method for DGNSS and INS integration with a window size of 10 seconds. The evaluation shows the advantage of CRT against the TC EKF estimator. However, the error sources for the pseudorange measurements are mitigated via double-difference before its integration with INS, leading to decimeter-level positioning accuracy. Therefore, the potential of CRT in challenging urban canyon using low-cost GNSS receiver, which can introduce large error in pseudorange error, is still to be explored. Continuous works are presented in [15] by adding more measurements [15] into the CRT. However, the performance of the CRT relies heavily on the size of the sliding window. The results showed that different window sizes led to different performance improvements. Effectively tuning the proper window size is one of the major factors for the performance of the CRT. The same phenomenon is also shown in our previous work in [17]. The too-large window size cannot guarantee the implementation performance and the too-small one cannot fully explore the time-correlation of the historical information. Therefore, how to effectively choose the window size is still an open question.

In short, the FGO-based GNSS/INS integration attracts numerous attention recently. Unfortunately, only limited evaluations are performed to partially show the effectiveness of the FGO, using simulated data [45] or highly accurate GNSS measurements from DGNSS [15]. In this paper, we go one step further where we make a comparison of the EKF and the FGO both in LC and TC of GNSS/INS integrations using the urban canyon dataset collected in Hong Kong, which is not done in the existing work [13, 15, 16, 45]. Moreover, we innovatively degenerate the FGO-based estimator to an "EKF like estimator" to show how the window size can affect the performance of FGO. Besides, we analyze the effects of window size against the performance of FGO and the reason behind it.

## III. Methodology

In this paper, four combinations of GNSS/INS are evaluated. 1) LC using EKF. 2) TC using EKF. 3) LC using the FGO. 4) TC using the FGO. We follow the reference in [46] and in [13] for the implementation of the EKF- and FGO-based integrations, respectively. The flowcharts of the implemented EKF and FGO ones are shown in Figures 1 and 2, respectively.

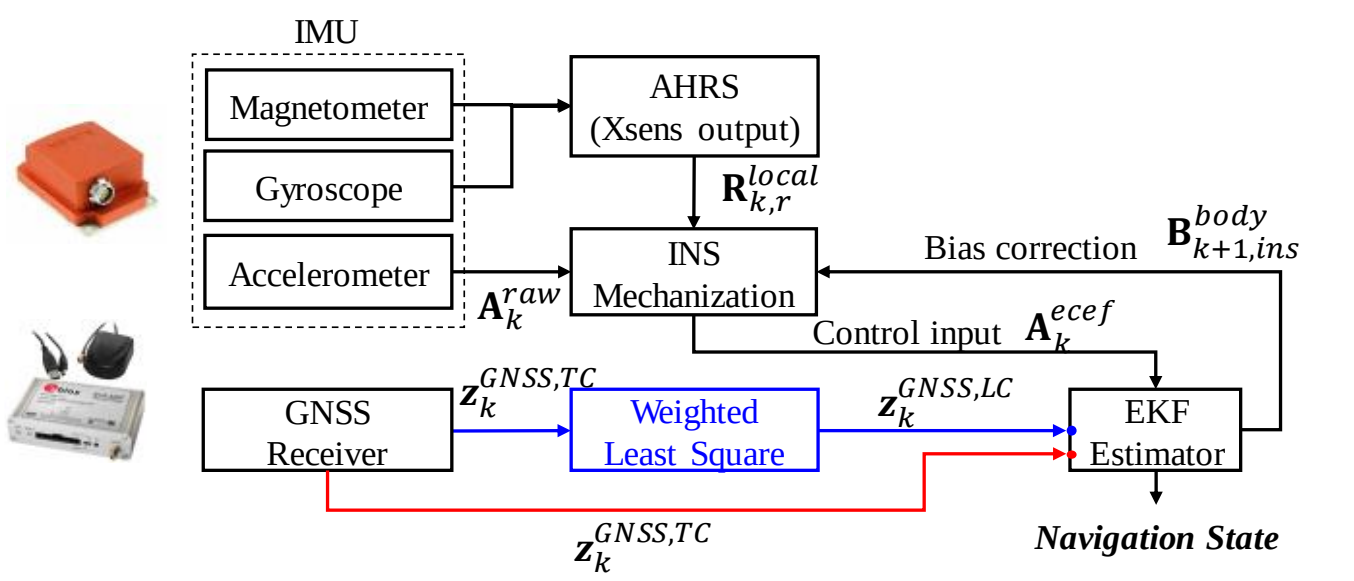

Fig. 1. The flowchart of the implemented LC (blue line) and TC (red line) GNSS/INS integrations using EKF, respectively.

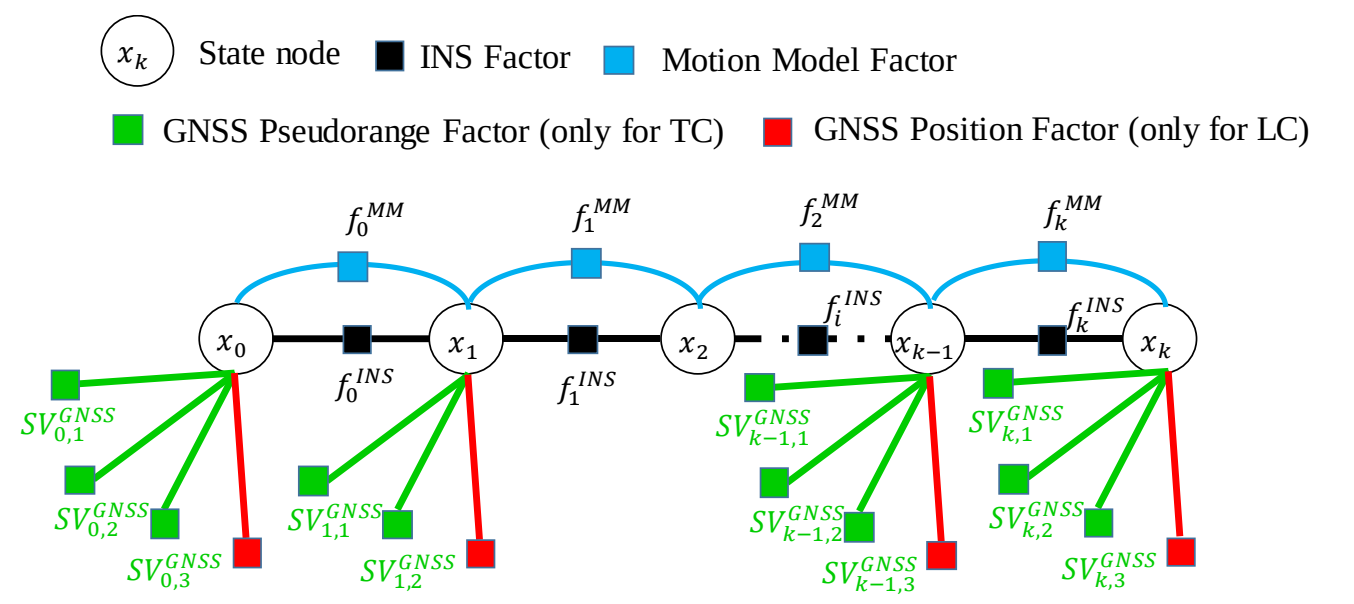

Fig.2 Illustration of the graph structure of the implemented LC and TC GNSS/INS integrations using FGO, respectively.

It is worth noting that only the positional performance is evaluated in this paper. The focus of this paper is to show how the FGO can help to mitigate the impacts of GNSS outlier measurements on GNSS/INS integration, compared with the EKF estimator. The estimated position state is in the ECEF frame. In this paper, we only use the linear acceleration measurements and attitude provided by an attitude and heading reference system (AHRS), which is a commercial solution of the INS we used. As for the GNSS, we use the raw pseudorange measurements for GNSS positioning. The methodologies of the four integrations are introduced as follows.

### A. Loosely-coupled GNSS/INS Integration Using EKF

The state-space of the system ($\mathbf{x}_k$) is represented as:

$$\mathbf{x}_k = (\mathbf{X}^{ecef}_{k,r}, \mathbf{V}^{ecef}_{k,r}, \mathbf{B}^{body}_{k,r})^T \quad (1)$$

where $\mathbf{X}^{ecef}_{k,r} = (x^{ecef}_{k,r}, y^{ecef}_{k,r}, z^{ecef}_{k,r})$ represents the position of the GNSS receiver in the ECEF coordinate (denoted by the subscript, $r$) [47] at a given epoch $k$. $\mathbf{V}^{ecef}_{k,r} = (vx^{ecef}_{k,r}, vy^{ecef}_{k,r}, vz^{ecef}_{k,r})$ denotes the velocities of the GNSS receiver. $\mathbf{B}^{body}_{k,ins} = (a^{body}_{k,x}, a^{body}_{k,y}, a^{body}_{k,z})$ denotes the bias of accelerometer in body (INS) frame. We use $\mathbf{R}^{local}_{k,r}$ to denote the attitude in the local frame provided by the AHRS.

The IMU measurements are expressed as follows:

$$\mathbf{A}^{raw}_{k} = (ax^{body}_{k}, ay^{body}_{k} az^{body}_{k})^T \quad (2)$$

where the $ax^{body}_{k}, ay^{body}_{k}, az^{body}_{k}$ represent the acceleration measurements in the body (INS) frame. As the estimated state $\mathbf{x}_k$ is in the global frame (ECEF), we need to transform the acceleration measurements from body frame to the global frame based on the attitude from AHRS. The transformed acceleration measurements in the global frame $\mathbf{A}_k^{ecef} = (ax_k^{ecef}, ay_k^{ecef}, az_k^{ecef})^T$ as follows [47]:

$$\mathbf{A}_k^{ecef} = \mathbf{R}_{GL}\mathbf{R}_{LB}(\mathbf{A}_k^{raw} - \mathbf{B}_{k,ins}^{body}) \quad (3)$$

The variable $\mathbf{R}_{LB}$ is the transformation matrix to transform the acceleration measurements from the body to the local frames based on $\mathbf{R}_{k,r}^{local}$ which can be expressed as follows:

$$\mathbf{R}_{LB} = \mathbf{R}_{LB}^{z}(\alpha)\mathbf{R}_{LB}^{y}(\beta)\mathbf{R}_{LB}^{x}(\gamma) \quad (4)$$

$$\text{with } \mathbf{R}_{LB}^{z}(\alpha) = \begin{bmatrix} \cos(\alpha) & -\sin(\alpha) & 0 \\ \sin(\alpha) & \cos(\alpha) & 0 \\ 0 & 0 & 1 \end{bmatrix}$$

$$\mathbf{R}_{LB}^{y}(\beta) = \begin{bmatrix} \cos(\beta) & 0 & \sin(\beta) \\ 0 & 1 & 0 \\ -\sin(\beta) & 0 & \cos(\beta) \end{bmatrix}$$

$$\mathbf{R}_{LB}^{x}(\gamma) = \begin{bmatrix} 1 & 0 & 0 \\ 0 & \cos(\gamma) & -\sin(\gamma) \\ 0 & \sin(\gamma) & \cos(\gamma) \end{bmatrix}$$

where the $\alpha$, $\beta$, and $\gamma$ denote the yaw, pitch, and roll angles, respectively. The variables $\mathbf{R}_{LB}^{z}(\alpha)$, $\mathbf{R}_{LB}^{y}(\beta)$ and $\mathbf{R}_{LB}^{x}(\gamma)$ denote the rotation matrixes corresponding to the yaw, pitch, and roll angles, respectively. $\mathbf{R}_{GL}$ is the transformation matrix to transform the acceleration measurement from local frame to the global frame based on the $\mathbf{x}_k$ with $\mathbf{R}_{GL}$ being expressed as follows:

$$\mathbf{R}_{GL} = \begin{bmatrix} -\sin(\emptyset_{lon}) & -\sin(\emptyset_{lat})\cos(\emptyset_{lon}) & \cos(\emptyset_{lat})\cos(\emptyset_{lon}) \\ \cos(\emptyset_{lon}) & -\sin(\emptyset_{lat})\sin(\emptyset_{lon}) & \cos(\emptyset_{lat})\sin(\emptyset_{lon}) \\ 0 & \cos(\emptyset_{lat}) & \sin(\emptyset_{lat}) \end{bmatrix} \quad (5)$$

where $\emptyset_{lon}$ and $\emptyset_{lat}$ represent the longitude and latitude based on the WGS84 geodetic system [47], which can be derived from $\mathbf{x}_k$.

The generic dynamic model of an LC EKF-based GNSS/INS integration can be written as:

$$\mathbf{x}_k = f(\mathbf{x}_{k-1}, \mathbf{u}_k) + \mathbf{w}_k \quad (6)$$

where $\mathbf{x}_{k-1}$ is the state in the previous epoch. $\mathbf{w}_k$ denotes the Gaussian noise associated with the $\mathbf{u}_k$, which represents the measurements from INS ($\mathbf{A}_k^{ecef}$ in this paper). Be noted that we only use the linear acceleration measurements in (6). The function $f(\mathbf{x}_{k-1}, \mathbf{u}_k)$ denotes the state transition function as follows:

$$f(\mathbf{x}_{k-1}, \mathbf{u}_k) = \begin{bmatrix} x_{k-1,r}^{ecef} + vx_{k-1,r}^{ecef} \cdot \Delta t \\ y_{k-1,r}^{ecef} + vy_{k-1,r}^{ecef} \cdot \Delta t \\ z_{k-1,r}^{ecef} + vz_{k-1,r}^{ecef} \cdot \Delta t \\ vx_{k-1,r}^{ecef} + ax_{k-1}^{ecef} \cdot \Delta t \\ vy_{k-1,r}^{ecef} + ay_{k-1}^{ecef} \cdot \Delta t \\ vz_{k-1,r}^{ecef} + az_{k-1}^{ecef} \cdot \Delta t \\ a_{k-1,x}^{body} \\ a_{k-1,y}^{body} \\ a_{k-1,z}^{body} \end{bmatrix} \quad (7)$$

where $\Delta t$ denotes the time difference between two epochs and the function $f(\mathbf{x}_{k-1}, \mathbf{u}_k)$ is based on the constant velocity model.

The measurements model of the LC EKF-based GNSS/INS integration can be written as:

$$\mathbf{z}_k^{GNSS,LC} = h^{GNSS,LC}(\mathbf{x}_k) + \mathbf{v}_k^{LC} \quad (8)$$

where $\mathbf{z}_k^{GNSS,LC}$ is the position measurements in ECEF frame from GNSS receiver represented as:

$$\mathbf{z}_k^{GNSS,LC} = (x_k^{GNSS}, y_k^{GNSS}, z_k^{GNSS})^T \quad (9)$$

The observation function $h^{GNSS,LC}(*)$ shows the relationship between the observation and the state at a $k$-th epoch as follows:

$$h^{GNSS,LC}(\mathbf{x}_k) = \begin{bmatrix} x_{k,r}^{ecef} \\ y_{k,r}^{ecef} \\ z_{k,r}^{ecef} \end{bmatrix} \quad (10)$$

$\mathbf{v}_k^{LC}$ is the Gaussian noise associated with the measurements and described with a covariance matrix equaling to $\mathbf{R}_k$. Regarding the calculation of $\mathbf{R}_k$, we follow the work in [48] which is calculated as follows:

$$\mathbf{R}_k = (h_{hdop} \cdot s_{UERE})^2 \mathbf{I}_{3x3} = \begin{bmatrix} (h_{hdop} \cdot s_{UERE})^2 & 0 & 0 \\ 0 & (h_{hdop} \cdot s_{UERE})^2 & 0 \\ 0 & 0 & (h_{hdop} \cdot s_{UERE})^2 \end{bmatrix} \quad (11)$$

where the $s_{USER}$ represents the user-equivalent range error (UERE) [48] which is set as 10 meters in this paper. $h_{hdop}$ denotes the horizontal dilution of precision (HDOP) [47] which can be calculated based on the GNSS measurements. $\mathbf{I}$ is a identity matrix with a size of $3 \times 3$.

### *B. Tightly-coupled GNSS/INS Integration Using EKF*

The flowchart is shown in Figure 2. The main difference between the LC and TC integrations are the domain of GNSS observations applied. The TC integration makes use of the raw GNSS pseudorange measurements. The state-space of the system ($\mathbf{x}_k$) is represented as:

$$\mathbf{x}_k = (\mathbf{X}_{k,r}^{ecef}, \mathbf{V}_{k,r}^{ecef}, \mathbf{B}_{k,r}^{body}, \boldsymbol{\delta}_{k,r}^{clock})^T \quad (12)$$

where the state is similar to the LC integration. The only difference is that the receiver clock bias $\boldsymbol{\delta}_{k,r}^{clock}$ also needs to be estimated. The measurement from INS and the dynamic models are identical to the ones in LC. The measurements model of the TC EKF-based GNSS/INS integration can be written as:

$$\mathbf{z}_k^{GNSS,TC} = h^{GNSS,TC}(\mathbf{x}_k) + \mathbf{v}_k^{TC} \tag{13}$$

where $\mathbf{z}_k^{GNSS,TC}$ is the GNSS pseudorange measurements ($N$ satellites in total) in ECEF frame from GNSS receiver represented as:

$$\mathbf{z}_k^{GNSS,TC} = (\rho_{k,1}^{GNSS}, \rho_{k,2}^{GNSS}, \dots, \rho_{k,i}^{GNSS}, \dots, \rho_{k,N}^{GNSS})^T \tag{14}$$

The $h^{GNSS,LC}(*)$ is the observation function which shows the relationship between the observation and the state at a $k$-th time instant. $\rho_{k,i}^{GNSS}$ represents the $i$th pseudorange measurement at epoch $k$. $N$ denotes the total number of satellites at epoch $k$. $\mathbf{v}_k^{TC}$ is the Gaussian noise associated with the measurements. The position of a satellite $\mathbf{SV}_{k,i}$ is represented as $\mathbf{x}_{SV,i}^{xyz} = (x_{SV}^{ecef}, y_{SV}^{ecef}, z_{SV}^{ecef})^T$. Therefore, we can obtain the predicted GNSS pseudorange measurement for satellite $\mathbf{SV}_{k,i}$ as:

$$h^p\left(\mathbf{SV}_{k,i}, \mathbf{X}_{k,r}^{ecef}, \boldsymbol{\delta}_{k,r}^{clock}\right) = ||\mathbf{x}_{SV,i}^{xyz} - \mathbf{X}_{k,r}^{ecef}|| + \boldsymbol{\delta}_{k,r}^{clock} \tag{15}$$

Therefore, the observation function $h^{GNSS,TC}(*)$ is formulated as follow:

$$h^{GNSS,TC}(*) = \begin{bmatrix} h^p\left(\mathbf{SV}_{k,1}, \mathbf{X}_{k,r}^{ecef}, \boldsymbol{\delta}_{k,r}^{clock}\right) \\ h^p\left(\mathbf{SV}_{k,1}, \mathbf{X}_{k,r}^{ecef}, \boldsymbol{\delta}_{k,r}^{clock}\right) \\ \cdots \\ h^p\left(\mathbf{SV}_{k,i}, \mathbf{X}_{k,r}^{ecef}, \boldsymbol{\delta}_{k,r}^{clock}\right) \\ \cdots \\ h^p\left(\mathbf{SV}_{k,N}, \mathbf{X}_{k,r}^{ecef}, \boldsymbol{\delta}_{k,r}^{clock}\right) \end{bmatrix} \tag{16}$$

Regarding the covariance matrix of $\mathbf{R}_k$ corresponding to the measurement vector $\mathbf{z}_k^{GNSS,TC}$, we follow the method in [49]. Each pseudorange measurement is given with different uncertainty based on its signal noise to the ratio (SNR) and satellite elevation angle. Given a satellite with SNR and elevation angle as $SNR_i$ and $el_i$, respectively. The weighting of the satellite is calculated as follows [49]:

$$W_i(el_i, SNR_i) == \frac{1}{sin^2 el_i}\left(10^{-\frac{(SNR_i - T)}{a}}\left(\left(\frac{A}{10^{-\frac{(F-T)}{a}}} - 1\right)\frac{(SNR_i - T)}{F-T} + 1\right)\right) \tag{17}$$

The parameter $T$ indicates the threshold of SNR. Parameters $a$, $A$, and $F$ are selected based on [49]. Therefore, the covariance matrix $\mathbf{R}_k$ is a diagonal matrix constituted by the weighting ${\sigma_i}^2$.

$$\mathbf{R}_k = \begin{bmatrix} {\sigma_1}^2 & \cdots & 0 \\ \vdots & \ddots & \vdots \\ 0 & \cdots & {\sigma_N}^2 \end{bmatrix} \tag{18}$$

$$\text{with } {\sigma_i}^2 = 1/W_i(el_i, SNR_i)$$

The subscript $N$ is the number of satellites and the matrix $\mathbf{R}_k$ is a $N \times N$ matrix.

### C. Loosely-coupled GNSS/INS Integration Using Factor Graph Optimization

In general, the goal of the multi-sensor integration is to find the optimal posterior state given the measurements from sensors. Therefore, the sensor integration problem can be formulated as a typical maximum a posteriori (MAP) problem [28]. In this paper, the measurements include two parts, GNSS and INS measurements. Assuming the GNSS and INS measurements are independent of each other, we can formulate the GNSS/INS integration as the following MAP problem:

$$\hat{\mathbf{x}} = \arg\max \prod_{k,i} P(\mathbf{z}_{k,i}|\mathbf{x}_k) \prod_k P(\mathbf{x}_k|\mathbf{x}_{k-1}, \mathbf{u}_k) \tag{19}$$

where the $\mathbf{z}_{k,i}$ represents the GNSS raw measurements at epoch $k$ and $\mathbf{x}_k$ represents the system state at epoch $k$. $i$ denotes the index of measurements at a given epoch $k$ (e.g. one epoch can have multiple pseudorange measurements). The variable $\mathbf{u}_k$ denotes the control input (e.g INS measurements) and $\hat{\mathbf{x}}$ is the optimal system state set [28]. The Bayes filter-based method finds the best estimation of the current state only considers: 1) the previous state. 2) control input and observation measurements at the current epoch. It fails to take full advantage of the historical information. Conversely, the FGO-based sensor integration [45] is applied to transfer the MAP problem into the non-linear optimization problem.

In the FGO-based integration, all the sensor measurements are treated as factors ($\zeta_j$) associated with specific states ($\mathbf{x}_j$). According to [50], the MAP problem is expressed as:

$$\widehat{\mathbf{X}} = \arg\max_{\mathbf{X}}\left(\prod_j \zeta_j(\mathbf{x}_j)\right) \tag{20}$$

$$\text{with } \zeta_j(\mathbf{x}_j) \propto \exp(-||h_j(\mathbf{x}_j) - \mathbf{z}_j||_{\Sigma_j}^2)$$

where $\zeta_j(\mathbf{x}_j)$ is a factor associated with a given measurements $\mathbf{z}_j$ which can be derived from both the GNSS and INS measurements. $\mathbf{x}_j$ is the state associated with the given measurements $\mathbf{z}_j$. $h_j(*)$ is the observation function associated with $\mathbf{z}_j$. The set $\mathbf{X} = \{\mathbf{x}_1, \mathbf{x}_2, \mathbf{x}_3, \dots, \mathbf{x}_k, \dots\}$ denotes the states that need to be estimated. Assuming that all the sensor noise is subject to Gaussian distribution, the negative logarithm of $\zeta_j(\mathbf{x}_j)$ is proportional to the error function [50] associated with measurements. Therefore, the formula (20) can be transformed as follows:

$$\widehat{\mathbf{X}} = \arg\min_{\mathbf{X}}\left(\sum_j ||h_j(\mathbf{x}_j) - \mathbf{z}_j||_{\Sigma_j}^2\right) \tag{21}$$

Therefore, the FGO transforms the (19) into a standard non-linear least squares (NLS) problem as (21) and obtain the optimal state set $\mathbf{X}$ by minimizing the derived error function.

The graph structure of the LC GNSS/INS integration using the FGO is shown in Figure 2. The state-space of the system is also represented as (1). The graph in Figure 2 includes all the historical observation measurements and states which is one of the major differences between the conventional Kalman filter-based [51] and the factor graph-based sensor integrations. The error functions of each listed factors are presented as follows.

*1) Motion Model Factor*

We use a constant velocity model to constraint the two consecutive states. Based on the constant velocity model, the motion model can be expressed as:

$$\mathbf{x}_k = h^{MM}(\mathbf{x}_{k-1}) + N(0, \mathbf{\Sigma}_k^{MM}) \tag{22}$$

where the $\mathbf{x}_k$ denotes the state at given epoch $k$. $h^{MM}(*)$ represents the motion model function as follows.

$$h^{MM}(\mathbf{x}_{k-1}) = \begin{bmatrix} x_{k-1,r}^{ecef} + vx_{k-1,r}^{ecef} \cdot \Delta t \\ y_{k-1,r}^{ecef} + vy_{k-1,r}^{ecef} \cdot \Delta t \\ z_{k-1,r}^{ecef} + vz_{k-1,r}^{ecef} \cdot \Delta t \\ \mathbf{B}_{k-1,r}^{body\,T} \end{bmatrix} \tag{23}$$

The $\mathbf{\Sigma}_k^{MM}$ is the covariance matrix associated with the motion model which is constant and is given based on the specification of the applied INS in this paper as follows:

$$\mathbf{\Sigma}_k^{MM} = \begin{bmatrix} 0.3^2 & 0 & 0 & 0 & 0 & 0 \\ 0 & 0.3^2 & 0 & 0 & 0 & 0 \\ 0 & 0 & 0.3^2 & 0 & 0 & 0 \\ 0 & 0 & 0 & 0.01^2 & 0 & 0 \\ 0 & 0 & 0 & 0 & 0.01^2 & 0 \\ 0 & 0 & 0 & 0 & 0 & 0.01^2 \end{bmatrix} \tag{24}$$

The units for the covariance matrix parts of $\mathbf{X}_{k,r}^{ecef}$ and $\mathbf{B}_{k,r}^{body}$ is given by meters and $m/s^2$, respectively. Therefore, the error function ($\mathbf{e}_k^{MM}$) of the motion model factor can be expressed as:

$$||\mathbf{e}_k^{MM}||_{\mathbf{\Sigma}_k^{MM}}^2 = ||\mathbf{x}_k - h^{MM}(\mathbf{x}_{k-1})||_{\mathbf{\Sigma}_k^{MM}}^2 \tag{25}$$

*2)* INS Factor

In LC FGO, the INS provides the linear accelerations which could directly correlate the velocities between two epochs. The acceleration measurements in the global frame are denoted as $\mathbf{A}_k^{ecef}$ (the $\mathbf{u}_k$) based on the formula (3). The measurement model for the linear acceleration is as follows:

$$\mathbf{x}_k = h^{INS}(\mathbf{x}_{k-1}, \mathbf{A}_k^{ecef}) + N(0, \mathbf{\Sigma}_k^{INS}) \tag{26}$$

with the measurement function $h^{INS}(\mathbf{x}_{k-1}, \mathbf{u}_k)$ as follows:

$$h^{INS}(\mathbf{x}_{k-1}, \mathbf{A}_k^{ecef}) = \begin{bmatrix} vx_{k-1,r}^{ecef} + ax_k^{ecef} \cdot \Delta t \\ vy_{k-1,r}^{ecef} + ay_k^{ecef} \cdot \Delta t \\ vz_{k-1,r}^{ecef} + az_k^{ecef} \cdot \Delta t \end{bmatrix} \tag{27}$$

where the covariance matrix for the INS factor is expressed as $\mathbf{\Sigma}_{k,acc}^{INS}$. Therefore, we can formulate the error function for INS acceleration measurements as follows:

$$||\mathbf{e}_k^{INS}||_{\mathbf{\Sigma}_k^{INS}}^2 = ||\mathbf{x}_k - h^{INS}(\mathbf{x}_{k-1}, \mathbf{A}_k^{ecef})||_{\mathbf{\Sigma}_k^{INS}}^2 \tag{28}$$

$\mathbf{\Sigma}_k^{INS}$ is constant and is given based on the specification of INS as follows:

$$\mathbf{\Sigma}_k^{INS} = \begin{bmatrix} 0.15^2 & 0 & 0 \\ 0 & 0.15^2 & 0 \\ 0 & 0 & 0.15^2 \end{bmatrix} \tag{29}$$

The units for the covariance matrix is given by $m/s$.

*3) GNSS Factor*

We can get the error function for a given GNSS measurement as follows for the LC FGO:

$$||\mathbf{e}_k^{GNSS}||_{\mathbf{\Sigma}_k^{GNSS}}^2 = ||\mathbf{z}_k^{GNSS} - h^{GNSS,LC}(\mathbf{X}_{k,r}^{ecef})||_{\mathbf{\Sigma}_k^{GNSS}}^2 \tag{30}$$

where $\mathbf{\Sigma}_k^{GNSS}$ denotes the covariance matrix which is calculated using the formulation (11).

### *D. Tightly-coupled GNSS/INS Integration Using Factor Graph*

The graph structure of the tightly-coupled GNSS/INS integration using the factor graph is shown in Figure 2. The state-space of the system is also represented according to (12). The motion and INS factors are the same as for the LC FGO,

*1) GNSS Pseudorange Factor*

The GNSS receiver can receive signals from multiple satellites at a given epoch *k* which can be expressed as:

$$\mathbf{SV}_k = \{\mathbf{SV}_{k,1}, \mathbf{SV}_{k,2}, \dots, \mathbf{SV}_{k,i}, \dots\ \mathbf{SV}_{k,N}\}, \tag{31}$$

Therefore, we can get the error function for a given satellite measurement $\rho_{SV,i}$ as follows:

$$||\mathbf{e}_{k,i}^{P}||_{\mathbf{\Sigma}_{k,i}^{SV}}^2 = ||\rho_{SV,i} - h^{GNSS,TC}(\mathbf{SV}_{k,i}, \mathbf{X}_{k,r}^{ecef}, \boldsymbol{\delta}_{k,r}^{clock})||_{\mathbf{\Sigma}_{k,i}^{SV}}^2 \tag{32}$$

where $\mathbf{\Sigma}_{k,i}^{SV}$ denotes the covariance matrix which is calculated using the formulation (18).

### E. Factor Graph Optimization

For the LC FGO, we formulate three kinds of factors including the motion model factor, the INS factor, and the GNSS factor. Therefore, the optimal state set $\mathbf{X} = \{\mathbf{x}_1, \mathbf{x}_2, \mathbf{x}_3, \dots, \mathbf{x}_k, \dots\}$ can be solved as follows:

$$\mathbf{X}^* = \operatorname{argmin} \sum_k ||\mathbf{e}_k^{GNSS}||_{\mathbf{\Sigma}_k^{GNSS}}^2 + ||\mathbf{e}_k^{MM}||_{\mathbf{\Sigma}_k^{MM}}^2 + ||\mathbf{e}_k^{INS}||_{\mathbf{\Sigma}_k^{INS}}^2 \tag{33}$$

For the TC FGO, we formulate three kinds of factors including the motion model factor, the INS factor, and the pseudorange factor. Therefore, the optimal state set $\mathbf{X} = \{\mathbf{x}_1, \mathbf{x}_2, \mathbf{x}_3, \dots, \mathbf{x}_k, \dots\}$ can be solved as follows:

$$\boldsymbol{X}^* = \operatorname{argmin} \sum_{i,k} ||\mathbf{e}_{k,i}^{P}||_{\mathbf{\Sigma}_{k,i}^{SV}}^2 + ||\mathbf{e}_k^{MM}||_{\mathbf{\Sigma}_k^{MM}}^2 + ||\mathbf{e}_k^{INS}||_{\mathbf{\Sigma}_k^{INS}}^2 \tag{34}$$

To solve the optimization problem, this paper makes use of the non-linear solver, GTSAM [52]. During the optimization, the Levenberg-Marquardt method is employed to solve the equation (33) and (34).

## IV. EXPERIMENT EVALUATION

### *A. Experiment Setup*

To evaluate the performance of the four listed integration schemes, we conduct experiments in an urban canyon of Hong Kong. The experimental vehicle and the tested scene is shown in Figure 3. The left-hand side of Figure 3 shows the experimental vehicle with all the sensors being installed in a compact sensor kit. The right figure shows the tested urban canyon in Hong Kong with tall buildings that are challenging for GNSS positioning.

During the test, a low-cost u-blox M8T GNSS receiver is used to collect raw GPS and BeiDou measurements at a frequency of 1 Hz. The Xsens Ti-10 IMU is employed to collect data at a frequency of 100 Hz. A fish-eye camera is employed to capture the sky view image to show environmental conditions, namely for analysis only. Besides, the NovAtel SPAN-CPT, a GNSS RTK/INS (fiber optic gyroscopes) integrated navigation system is used to provide the ground truth. The gyro bias in-run stability of the FOG is 1 degree per hour and its random walk is 0.067 degree per hour. The baseline between the rover and GNSS base station is about 7 km. All the data are collected and synchronized using the robot operating system (ROS) [53]. The coordinate systems between all the sensors are calibrated before the experiments. We run both the EKF and FGO using a high-performance desktop computer with an Intel i7-9700K at 4.20GHz and 64GB RAM. The applied parameters in this paper are shown in Table 1. We are aware that the choice of covariance can significantly affect the accuracy of GNSS/INS. Therefore, we use the same covariance parameters for EKF and FGO based on (11) and (18).

As the performance comparison of GNSS/INS integration using EKF has been extensively conducted in the existing research [36], we focus on analyzing the difference between the EKF and the FGO. The estimated state is in the Earth-centered Earth-fixed (ECEF) frame [47]. We transform the positioning results from the ECEF into an east, north, and up (ENU) frame [47]. As the orientation is directly from the AHRS, only the 2D positioning (north and east directions) accuracy is evaluated.

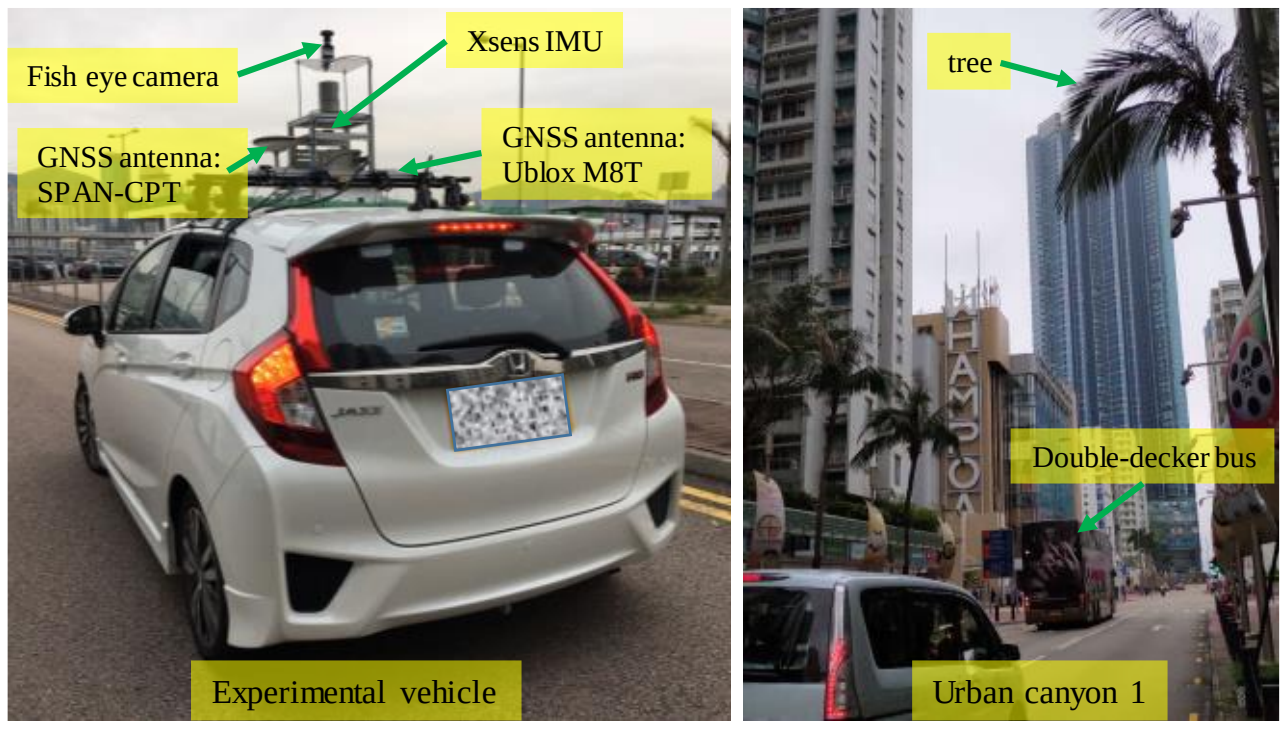

Fig.3 Left figure shows the experimental vehicle and sensors setup. The right figure illustrates the tested urban canyon in Hong Kong.

## B. *Comparison of Positioning Accuracy*

The positioning performances of the integrations in the tested urban canyon are shown in Table 2. 9.14 meters of 2D mean error is obtained using LC EKF with a standard deviation of 7.60 meters. The error decreases to 8.03 meters after using the TC EKF method. Moreover, the time efficiency of the loosely and tightly EKF is similar to less than 0.1 seconds being spent to process all the data. After applying the LC FGO to integrate the GNSS/INS, the 2D error decreases to 7.01 meters with a standard deviation of 6.41 meters, which is even better than that of the TC EKF. Moreover, the standard deviation is also reduced from 7.15 to 6.41 meters. However, 1.52 seconds is consumed to process all the data. Interestingly, the 2D mean error decreases to 3.64 meters after using the TC FGO. The standard deviation also decreases dramatically from 6.41 to 2.84 meters compared with that of the LC FGO. However, the computational load increases by two times due to the increased number of factors in TC FGO, compared with the LC one. In short, the best performance is obtained using the TC FGO. As all the historical data are considered in the FGO, a higher computation load is caused accordingly.

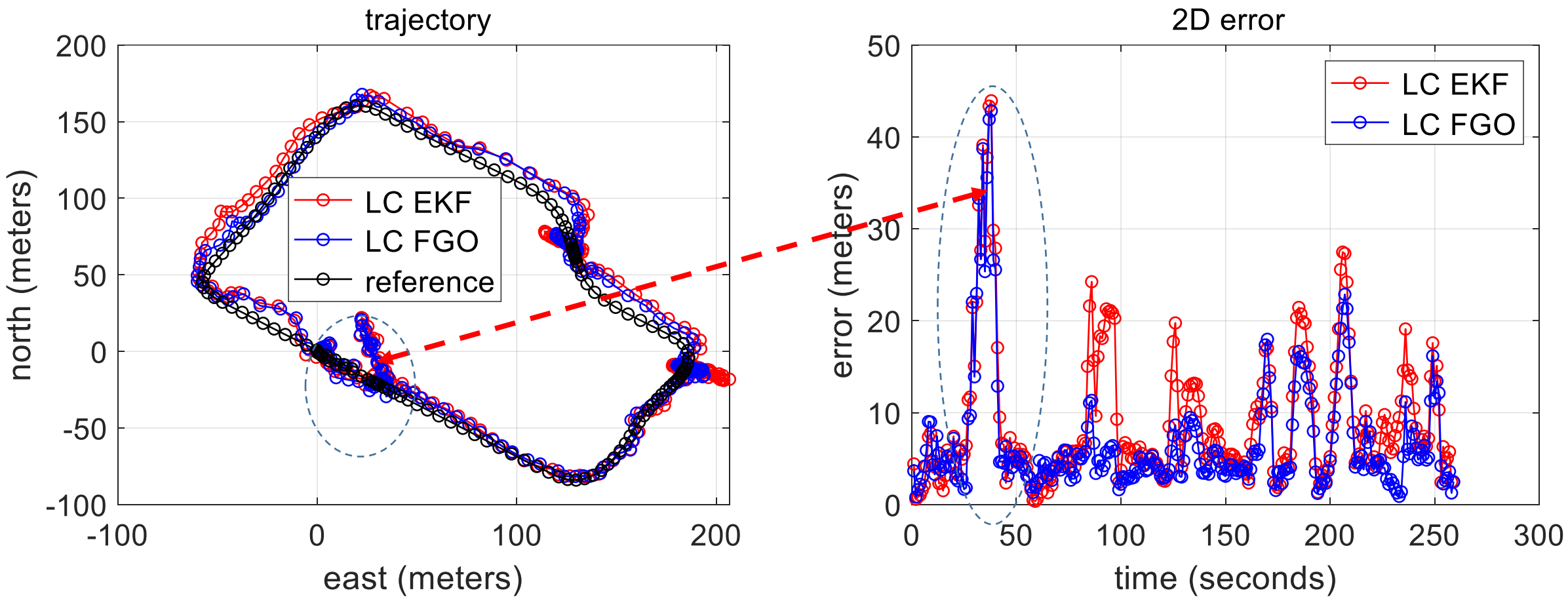

Fig.4 Trajectories of the LC GNSS/INS integrations using EKF and FGO in the east, north, and up (ENU) frame. The black curve denotes the reference trajectory. The red and blue curves in the left-hand side figure represent the trajectories from LC integrations using EKF and FGO, respectively. The right figure shows the 2D errors during the test.

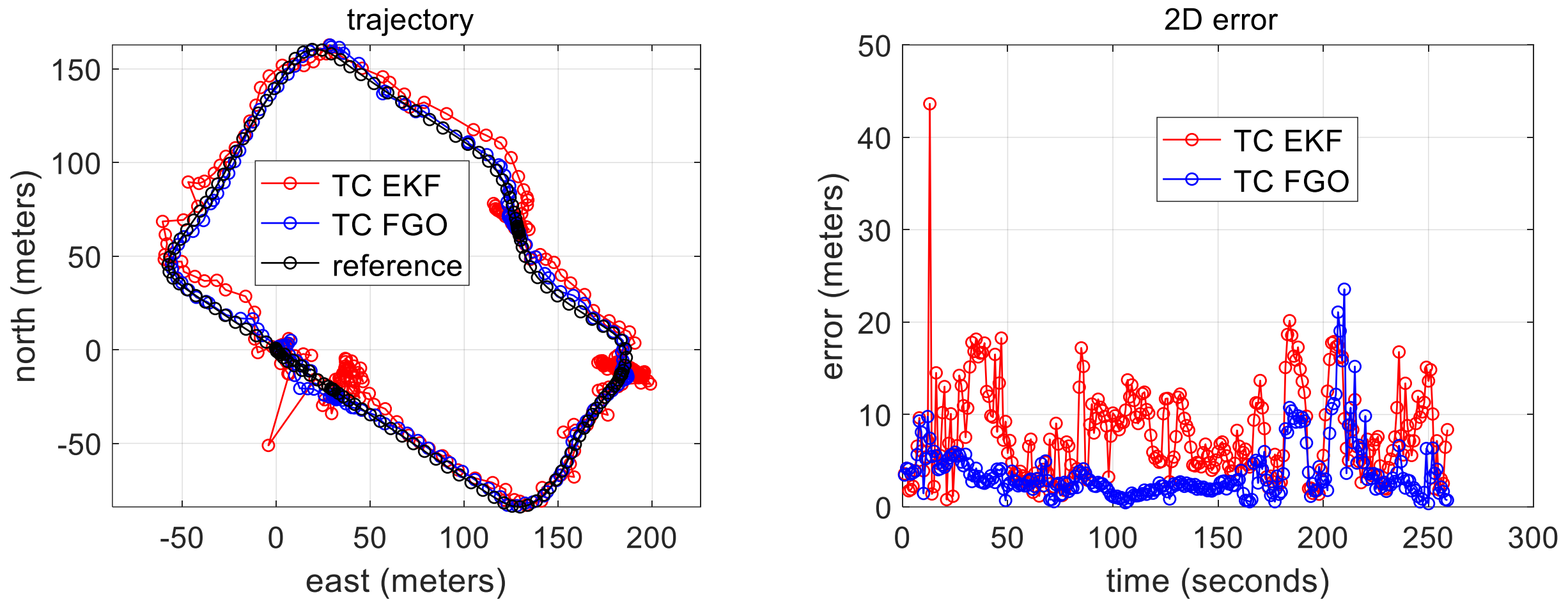

Fig.5 Trajectories of TC GNSS/INS integrations using EKF and FGO in the east, north, and up (ENU) frame. The black curve denotes the reference trajectory. The red and blue curves in the left-hand side figure represent the trajectories from TC integrations using EKF and FGO, respectively. The right figure shows the 2D errors during the test.

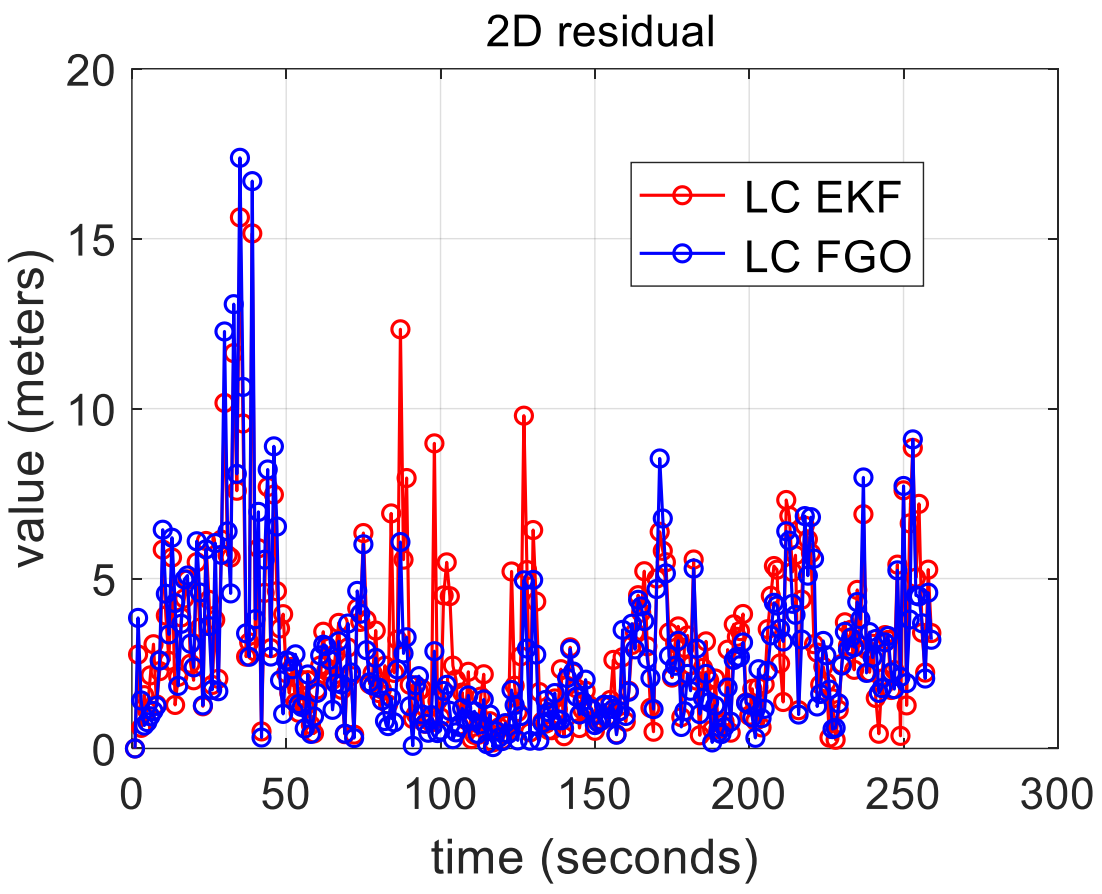

Fig.6 2D residuals of the loosely-coupled GNSS/INS integrations using EKF and FGO.

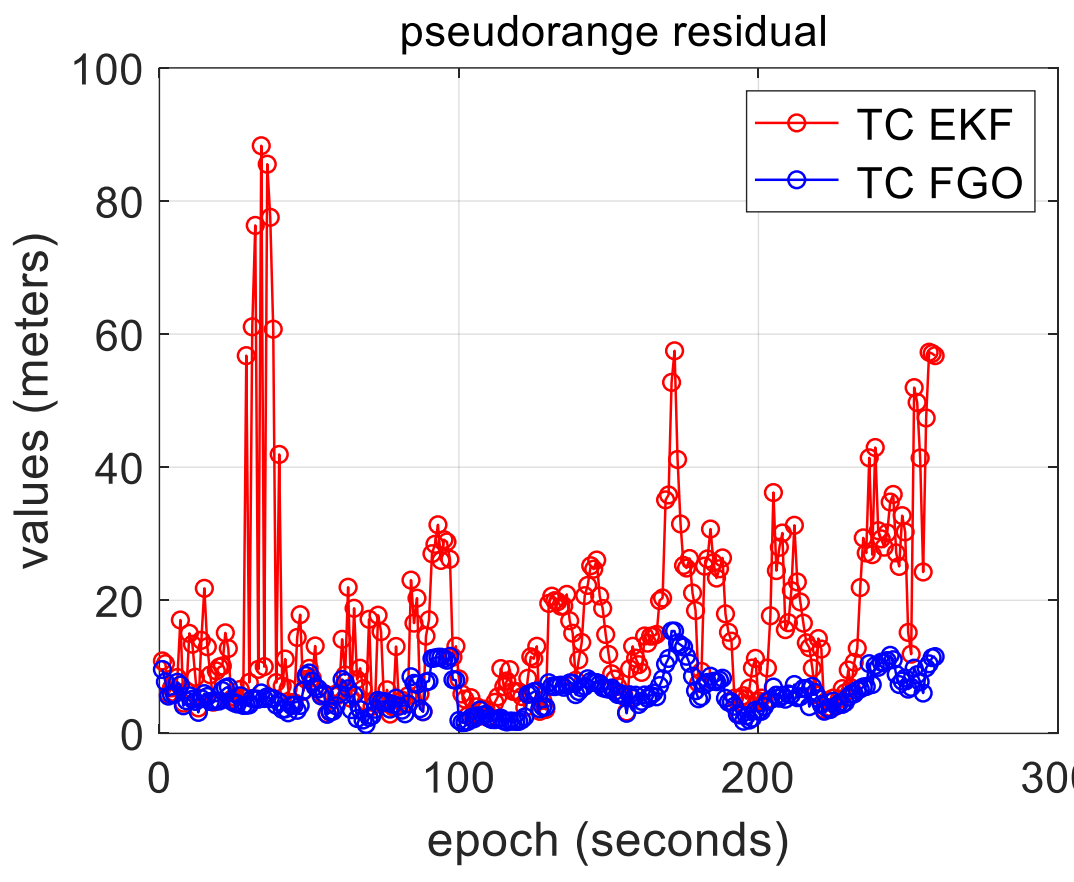

Fig.7 2D residuals of the loosely-coupled GNSS/INS integrations using EKF and FGO.

**Table.1 Parameter values used in this paper**

| Parameters | $F$ | $T$ | A |
|---|---|---|---|
| Values | 10 | 45 | 30 |
| Parameters | $s_{UERE}$ | A | |
| Values | 10 m | 32 | |

**Table.2 Positioning Performance and computational load (used time) of the four methods**

| All data | Loosely EKF | Tightly EKF | Loosely FGO | Tightly FGO |
|---|---|---|---|---|
| **Mean error** | 9.14 m | 8.03 m | 7.01 m | 3.64 m |
| **Std** | 7.60 m | 7.15 m | 6.41 m | 2.84 m |
| **Used Time** | 0.053 s | 0.071 s | 1.52 s | 3.74 s |

The trajectories and the positioning error of the two LC and two TC integrations during the test are shown in Figures 4 and 5, respectively. We can see from the right-hand side of Figure 4, the mean error of LC FGO is slightly improved in most of the epochs compared with the LC EKF. However, the improvement is still limited. However, the mean error is significantly improved after using the TC FGO which can be seen from the right-hand side of Figure 5.

The residual is a term evaluating the difference between the measurements and optimal state estimation. If all the applied measurements are perfectly accurate, the residual should be zero. However, the remaining residuals are usually not zero due to the noise caused by signal blockage or reflection, leading to the NLOS receptions in GNSS pseudorange measurements. The objectives of both the EKF- and FGO-based methods are to minimize the residuals of all the considered measurements based on the associated covariance matrix. Smaller residuals usually mean that the estimated state is closer to the optimal estimation. In short, the residual is a decent indicator of the quality of the GNSS/INS integration. Therefore, we also present the residual results of the four listed combinations.

The residuals of the LC and TC integrations are shown in Figures 6 and 7, respectively. For the LC integration, the observation is the measurement of GNSS positioning. The residual ($p_{r,LC}$) denotes the difference between the GNSS positioning and the final GNSS/INS integrated result which can be calculated as follows.

$$p_{r,LC} = ||\mathbf{z}_k^{GNSS,LC} - h^{GNSS,LC}(\mathbf{x}_k{}^*)|| \quad (34)$$

where $p_{r,LC}$ denotes the residual of GNSS in LC integration, the $\mathbf{x}_k{}^*$ denotes the state estimation at a given epoch *k*. We can see from Figure 6 that the FGO-based method has a similar or smaller residual compared with that the EKF-based integration throughout the test.

The pseudorange residual in the TC integrations can be calculated as follows:

$$\rho_{r,TC} = 1/N \sum_{i=1}^{N} (\rho_{k,i}^{GNSS} - h^{GNSS,TC}(\mathbf{SV}_{k,i}, \mathbf{X}_{k,r}^{ecef}, \boldsymbol{\delta}_{k,r}^{clock})) \quad (35)$$

where the $\rho_{r,TC}$ denotes the residual of GNSS in TC integration at a given epoch *k*. As shown in Figure 7, the FGO-based method possesses a significantly smaller residual. Be noted that the residual for TC integration is in the GNSS pseudorange domain. The small residual means that the final positioning result is closer to the given measurement. We can also see from Figure 7 that the residual of TC FGO is significantly smoother, compared with that of the TC EKF. In short, the FGO-based GNSS/INS integration obtains better accuracy compared with the EKF-based one. Moreover, the TC integration using the FGO obtains the best performance among the four listed integrations.

### *C. Analysis of the Improvement from Factor Graph Optimization Vs. Window Size*

The GNSS/INS integration results based on tested data in this paper show that both the LC [13] and TC [45] integrations obtain better performance using FGO, compared with the EKF-based integration. However, the improvement from the LC integration using FGO is limited, compared with the EKF one. This is because the LC integration cannot effectively model the uncertainty of the raw GNSS pseudorange measurements. In contrast, the significant improvement from TC integration using FGO motivates us to find out the reason behind it. According to [10], one of the major differences between the EKF and FGO is the "iterations" applied. The EKF-based sensor integration only iterates once based on the given observation measurements. However, the FGO iterates several times based on all the historical and current measurements to approach the optimal state.

Inspired by this, we propose to degenerate the FGO into an "EKF like estimator" in the TC GNSS/INS integrations to further validate the contribution of iterations in FGO. Be noted that the TC GNSS/INS integrations have larger amounts of factors, compared with the LC FGO. In other words, more constraints are considered during optimization. In fact, according to (33), if the window size of the optimization is set to 1 second, meaning that the TC integrations using FGO only considers the measurements at the current and last epochs, the major difference between the FGO and the EKF is the number of iterations during the GNSS/INS integrations. The window size denotes the epoch of historical states considered in the FGO. We call the FGO with a window size of 1 second as the "EKF like estimator". Figure 8 shows the 2D positioning error of TC GNSS/INS integration using FGO under different window sizes. The 2D mean positioning error is 5.18 meters (black curve in Figure 8) when the window size is 1 second which is still better than the TC integration using EKF (8.03 meters). As the major difference between the "EKF like estimator" and the EKF is the number of iterations. Therefore, this improvement (from 8.03 to 5.18 meters) is mainly contributed by the iterations compared with the EKF-based integrations. The result shows that the multiple iterations in FGO are one of the reasons behind the improvement. As the FGO is a process of finding the optimal estimation based on the gradient [10], multiple iterations can effectively help to find the optimal state estimation.

According to (33) and the optimization process, the re-linearization is performed at each iteration based on all the considered measurements. However, the EKF-based integration only performs the linearization once based on the predicted state at the current epoch. Meanwhile, the accuracy of the linearization relies heavily on the accuracy of the linearization point. The re-linearization is the other reason causing which facilitates the improvement of performance in FGO, compared with the EKF. Moreover, the re-linearization is conducted concerning all the states which can also enhance the robustness of the FGO against the outlier measurements. This phenomenon can be seen in Figure 8 that the green curve (window size is 5s) peaks near epoch 35 with the 2D error reaching more than 50 meters. However, with the increased window size, the 2D error is significantly reduced and the smoother result is obtained.

In short, we argue that the improvements from the FGO arise from the following aspects:

(1) **Multiple Iterations**: The multiple iterations effectively help the FGO to approach the optimal state estimation. In the TC GNSS/INS integration, the observation function for pseudorange is a non-linear, and a single iteration in EKF is hard to reach optimal.

(2) **Re-linearization**: In each iteration of FGO, the linearization is conducted again. As the observation function for pseudorange is highly non-linear, multiple linearizations can effectively mitigate the linearization error. Therefore, multiple iterations and re-linearization are complementary.

(3) **Time-correlation**: The FGO effectively considers the historical information and all the historical information is connected by the INS factor. As a result, the time-correlation between epochs are explored and used to resist the outliers. In other words, the proper window size of FGO can help to improve the performance. The same finding is also discussed in the work of [15]. However, the findings are not validated with a real dataset. The following of this section focuses on analyzing the effects of window size against the performance of FGO, by considering both environmental conditions and the GNSS pseudorange error distribution.

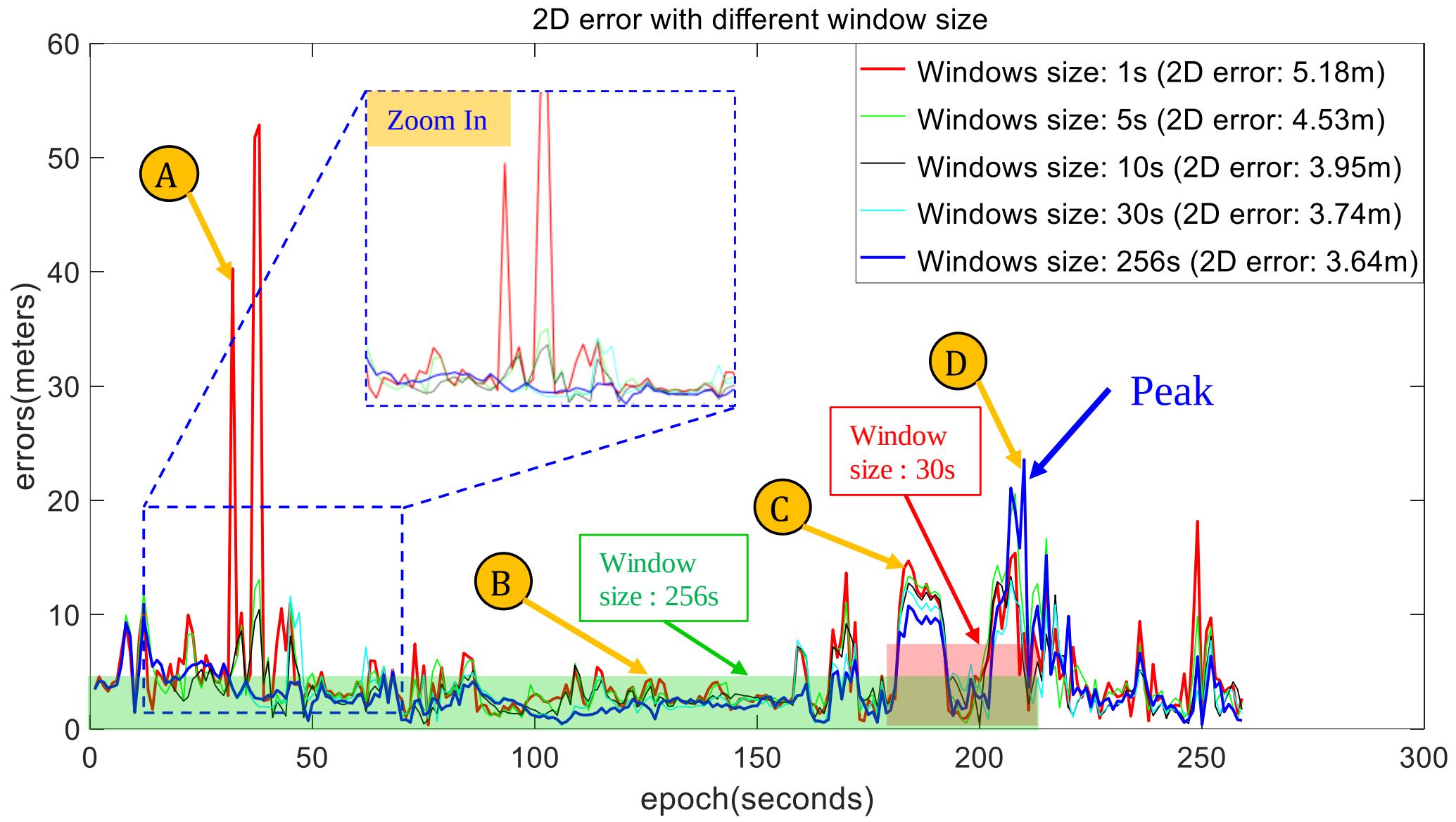

Fig.8 2D positioning errors under different window sizes used in TC GNSS/INS integrations using FGO. The x-axis denotes the epochs and the y-axis represents the value of 2D positioning errors. The shaded area using red and green rectangles denote the sliding windows with a size of 30 and 256 seconds, respectively.

According to the work in [10], the other major difference between the EKF and the FGO is that more historical data is considered during the optimization in FGO. To see the effects of window size on the performance of GNSS/INS integration using factor graph, we present the positioning results using different window sizes which are shown in Figure 8. Be noted that the window size of 256s equals batch optimization, which considers all the historical information in FGO. When the window size is 1 second, the improvement from the FGO is limited. With the increased window sizes (5 s, 10 s, 30 s, 256 s), more historical information is considered during the optimization and the overall performances are improved gradually. Besides, the errors (blue curve) arising from the batch optimization is significantly smoother than the other curves. Overall, more historical data tend to enhance the resilience against outliers in GNSS measurements, such as the NLOS receptions and multipath effects. Interestingly, we can find that when the window size is about 30s, the accuracy (3.74 meters) is close to the one (3.64 meters) from batch optimization (window size is 256 s).

Interestingly, we can see from Figure 8, the error (blue curve) of batch optimization at epoch D is conversely larger than the results arising from smaller window size (e.g. the 20s, 10s, and 5s). This means that the larger window size does not necessarily lead to better performance in FGO. Figure 9 shows the sky view images captured by a fish-eye camera and satellite visibilities for the four selected epochs (epochs A, B, C, and D) in Figure 8. The blue and red circles denote the LOS and NLOS satellites, respectively. The LOS and NLOS satellites are classified based on our previous work in [14]. The errors peak at epochs A, C, and D due to the severer NLOS receptions (see red circles). The period near epoch B introduces a similar positioning error (3~6 meters) using different window sizes. In short, the actual sensor noise (error magnitude) model near epoch B is significantly different from the one near epoch D. As a result, the too old historical measurements cannot reflect the measurements noise at current epoch D. The larger window size (see blue curve in Figure 8) of FGO even leads to worse result at the epoch D.

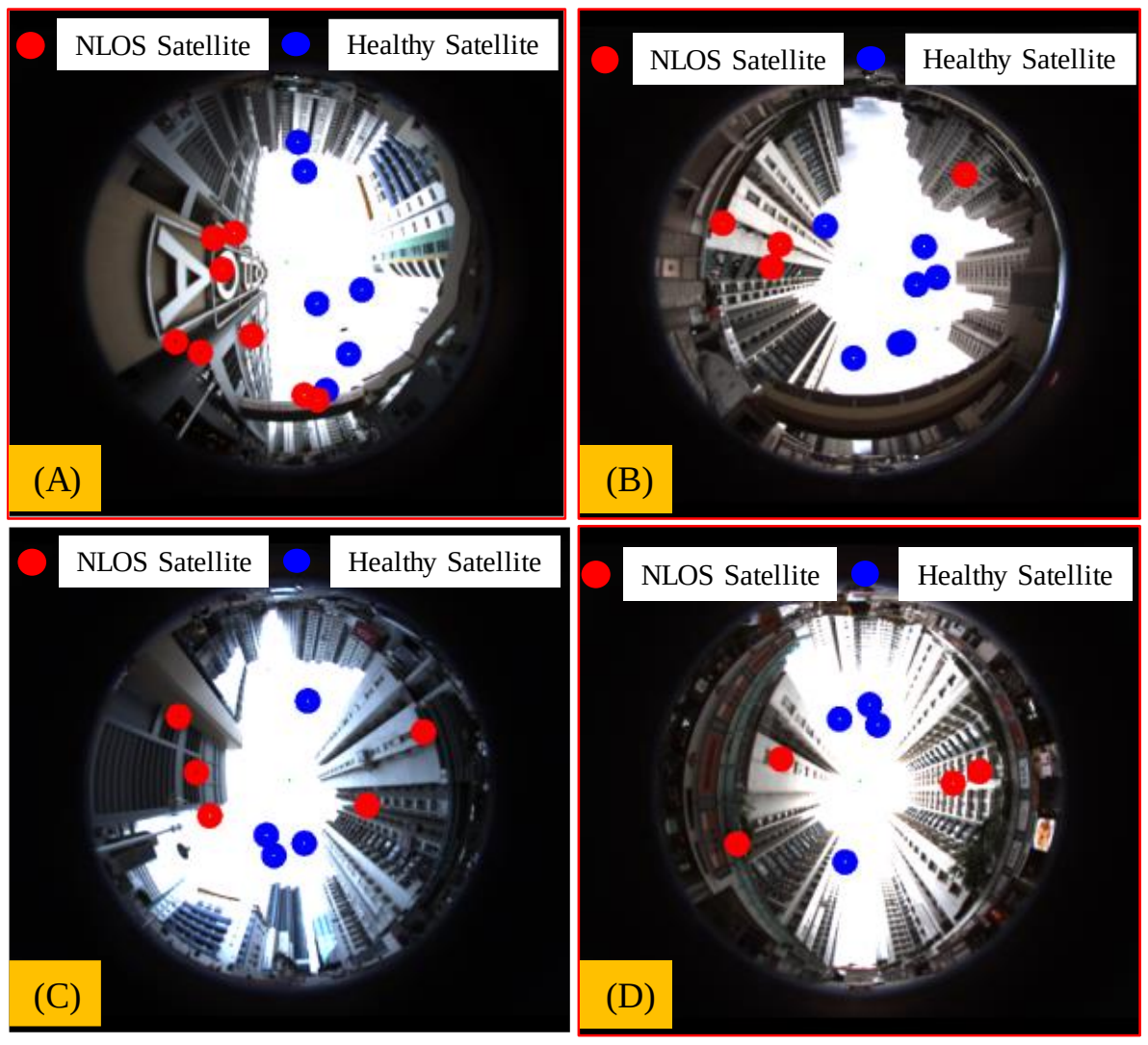

Fig.9 Skyview and satellite visibilities of the four selected epochs in Figure 8. The blue and red circles denote the LOS and NLOS satellites, respectively.

Both the EKF and FGO-based GNSS/INS integrations rely on the assumption of the Gaussian model regarding the sensor noise. Unfortunately, this assumption is usually violated due to the satellite signal reflection and blockage from buildings. This is one of the major factors limiting the performance of GNSS/INS integration in urban canyons. Instead of using Gaussian distribution, recently, the team from Chemnitz University of Technology [38] proposes to make use of the Gaussian mixture model to model the pseudorange measurements noise. They found that if the residuals of all the measurements inside a window can effectively model the error distribution of the measurements at the current epoch, the FGO can be significantly improved based on an estimated Gaussian mixture model (GMM) using historical pseudorange residuals. The work in [38] argues that the noise of pseudorange measurements is not subjected to Gaussian distribution in urban canyons. Inspired by their work, we show the pseudorange errors and residuals distributions near epoch D using different window sizes.

Figure 10 shows the exact GPS/BeiDou pseudorange errors which are labeled based on the ground truth trajectory and 3D building modeling using the ray-tracing technique [54]. In other words, Figure 10 shows the true error distribution of pseudorange measurements. Be noted that the pseudorange errors are within a window of epoch D with a size of 30 s which is shown in the shaded area by the red rectangle in Figure 8. We can see from the histograms of Figure 10 that there are roughly three peaks. This coincides with the finding in [54] that three Gaussian components are usually enough to model the pseudorange error distribution in urban canyons. Interestingly, we can see that the histogram has a long tail on the right-hand side. This is mainly arising from NLOS receptions, caused by building reflections. This is one of the main factors resulting in the Non-Gaussian property in pseudorange measurements. A similar long-tail phenomenon is also witnessed in the BeiDou pseudorange errors on the right-hand side of Figure 10. Similar to [54], we use a GMM with three Gaussian components to quantitatively fit the GPS/BeiDou error distributions and the GMM are shown by the blue curves. The mean, standard deviation, and weighting of each Gaussian component are shown in the figure. The first component of the GMM for GPS pseudorange measurements with a mean of 38.76 meters represents the NLOS signals which introduce a long tail phenomenon. The first component of the GMM for BeiDou pseudorange measurements with a mean of 32.62 meters represents the NLOS signals as well. In short, we can see that the numerous NLOS receptions leads to severe long-tail in the error distribution.

Similar to Figure 10, Figure 11 shows the residuals of GPS/BeiDou pseudorange measurements which are calculated based on (35). Interestingly, similar histograms and GMM parameters (see the table inside Figure 11) are obtained using the pseudorange residuals. This means that the fitted GMM in Figure 11 using the pseudorange residuals can effectively represent the actual error distribution (the GMM shown in Figure 10). As Figure 8 shows, a window size of 30 s leads to similar accuracy, compared with the batch optimization. This means that the historical measurements within the 30 s window have the potential to describe the error distribution of the measurement noise near epoch D. This coincides with the results shown in Figures 10 and 11.

Figure 12 shows the distributions of the residuals with a significantly larger window size of 256 seconds, compared with Figures 10. We can see from the left-hand side of Figure 12; the GPS residuals introduce a long-tail with the maximum pseudorange residual reaching 100 meters which is significantly different from the one from Figure 11. As a result, the Gaussian component with a mean of 36.58 introduces a large standard deviation of 1100.6 meters. The mean values (10.07 meters) of the GMM for the BeiDou pseudorange measurement are significantly reduced compared with the 32.62 meters shown on the right-hand side of Figure 10. In short, the residuals within the window size of 256 seconds at epoch D cannot effectively describe the actual noise of epoch D (see Figure 10). We believe that this is a significant factor for the optimal window size determination of FGO.

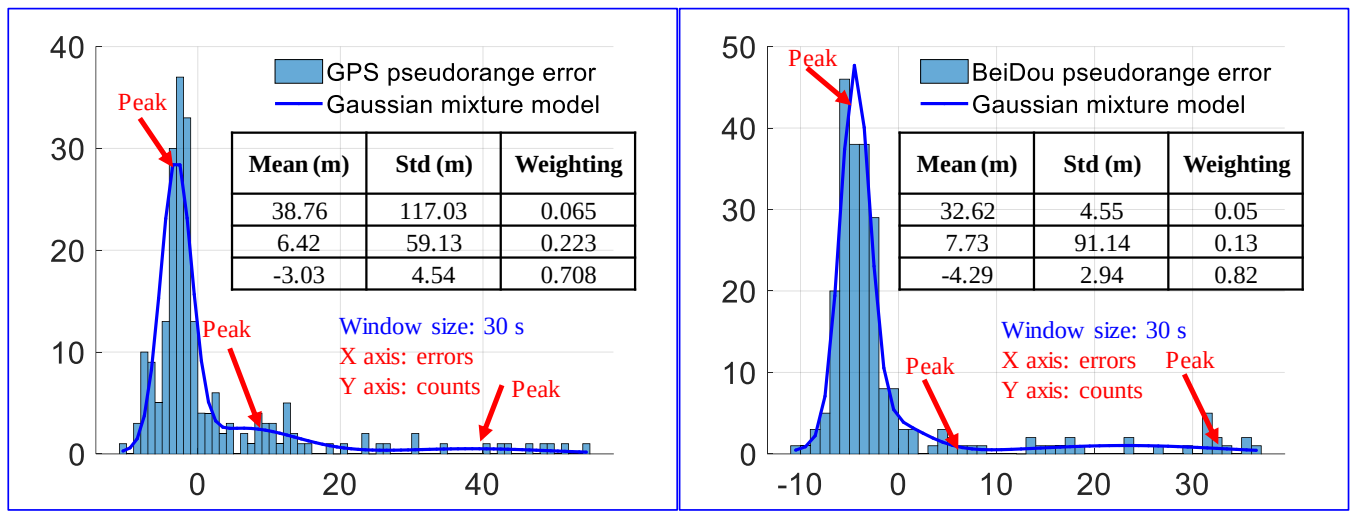

Fig.10 Histogram of GPS (left) and BeiDou (right) pseudorange errors and respected fitted GMM inside a sliding window of epoch D with a size of 30 seconds (see the shaded area by the red rectangle in Figure 8). The x-axis denotes the value of pseudorange errors acquired by the ray-tracing technique. The y-axis denotes the counts of errors within hist. The blue curve represents the GMM fitted by the histogram using 3 Gaussian components.

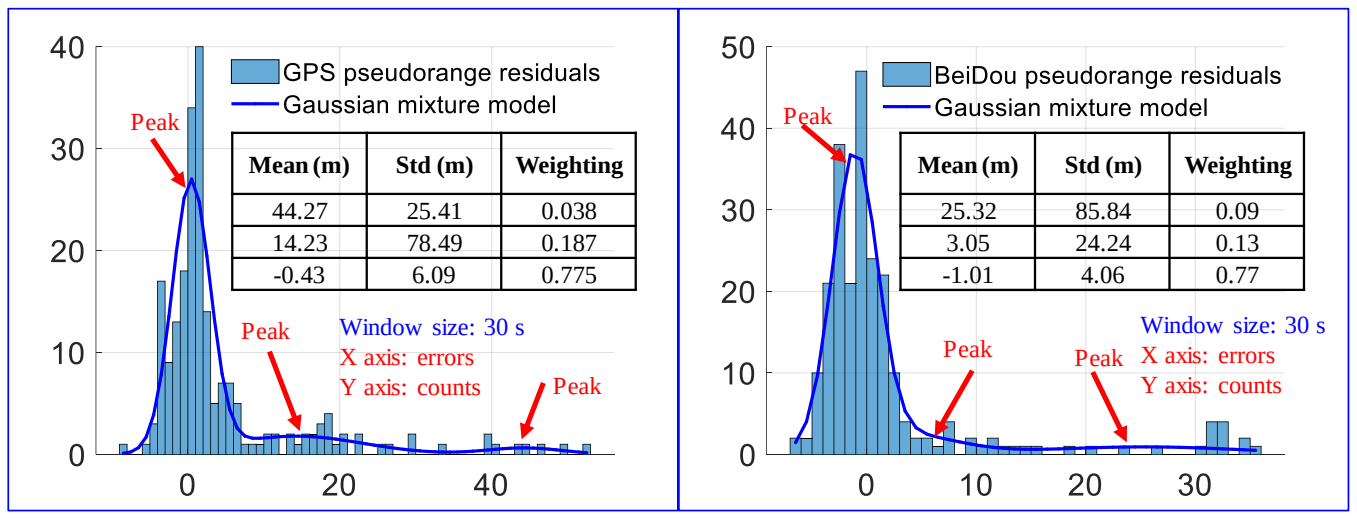

Fig.11 The histogram and GMMs of pseudorange residuals near epoch D with a sliding window of 30 s. Similar to Figure 10. The major difference is that the histogram and GMMs are based on residuals (see (35)).

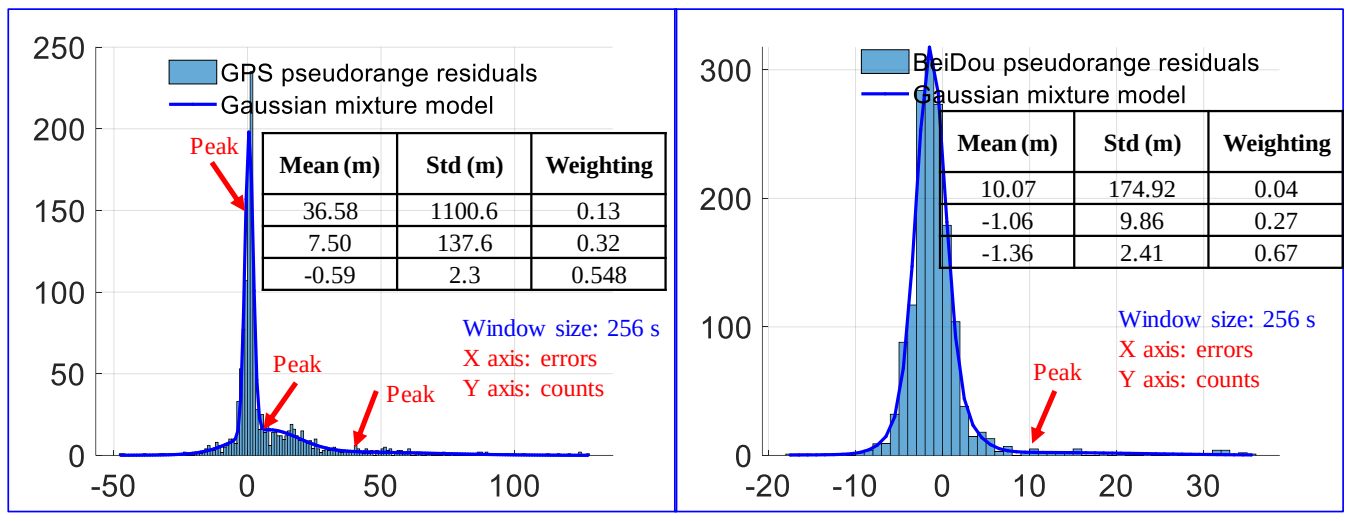

Fig.12 Similar to Figure 11. The histogram and GMMs of residuals near epoch D with a sliding window of 256 s.

## V. Conclusions And Future Work

GNSS/INS integration is significant for autonomous systems with navigation requirements. This paper comprehensively compares four GNSS/INS integrations using both EKF and FGO with the real dataset collected in urban canyons. More importantly, we analyze the effects of window size against the performance of FGO based on the validated dataset, by considering both the GNSS pseudorange error distribution and environmental conditions. Firstly, the frameworks and corresponding state functions of the four integrations are presented. Then the real road test is conducted to evaluate the corresponding performances. The experiment results show that the TC GNSS/INS integration can obtain better performance, compared with that of the LC one. The TC integration using FGO obtains the best performance in the comparison. According to the analysis, the superior performance of FGO compared with EKF is caused by three parts: 1) the multiple iterations; 2) the larger size of data applied in the optimization. 3) the re-linearization against all the states. We believe that the FGO-based sensor fusion will be the promising replacement of the EKF in the coming decades. The analysis of the window size against the performance of FGO shows that the proper window size is highly correlated with the environmental conditions. Using environmental context-awareness to identify the sensor noise characteristics could be a promising solution to adaptively tune the window size.

As the evaluation of the paper is limited to one dataset collected in a typical urban canyon of Hong Kong and analyzes the reasons behind the results, how the FGO will work using diverse scenarios urban scenarios is interesting to see. We will present the performance of FGO using more sensors (e.g. LiDAR) using our recently published UrbanLoco dataset [55] which involves the full suit sensor data for vehicular navigation collected in both Hong Kong and downtown San Francisco. Besides, the study of the appropriate window size tuning for the FGO will also be conducted in future work.

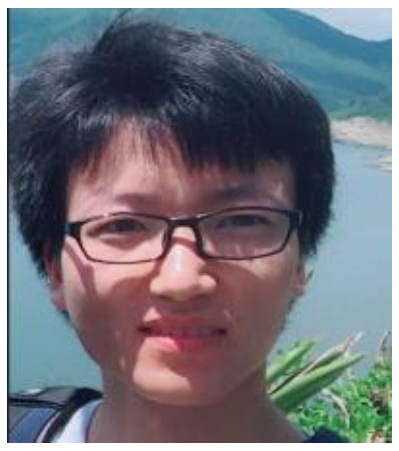

**Weisong Wen** was born in Ganzhou, Jiangxi, China. He is a Ph.D. candidate in mechanical engineering, the Hong Kong Polytechnic University. His research interests include multi-sensor integrated localization for autonomous vehicles, SLAM, and GNSS positioning. He was a visiting student researcher at the University of California, Berkeley (UCB) in 2018.

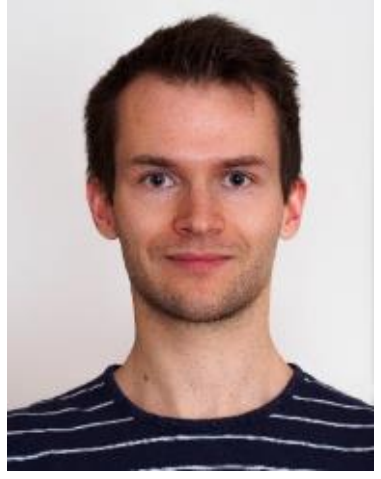

**Tim Pfeifer** is a Ph.D. candidate at the Chemnitz University of Technology, Germany. His research interests include factor graphs, robust estimation techniques, and Gaussian mixture models for robot localization. He focuses on methods to make autonomous systems robust against unforeseen sensor events.

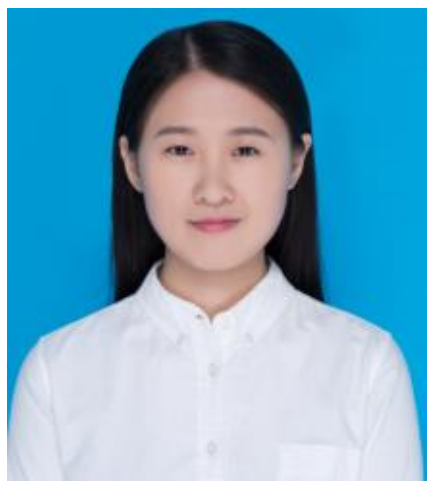

**Xiwei Bai** is currently a research assistant with the Interdisciplinary Division of Aeronautical and Aviation Engineering, at Hong Kong Polytechnic University. Her research topic is GNSS positioning aided by computer vision, object detection, and recognition using the deep neural network.

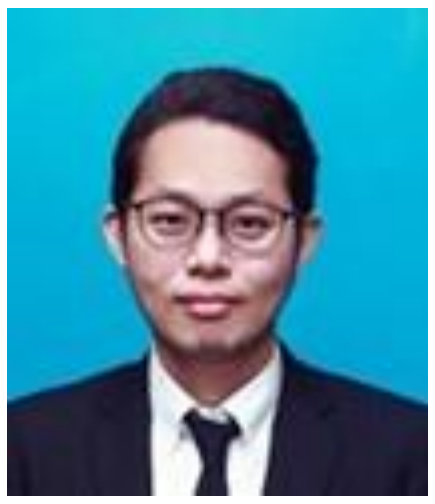

**Li-Ta Hsu** received the B.S. and Ph.D. degrees in aeronautics and astronautics from National Cheng Kung University, Taiwan, in 2007 and 2013, respectively. He is currently an assistant professor with the Interdisciplinary Division of Aeronautical and Aviation Engineering, The Hong Kong Polytechnic University, before he served as a post-doctoral researcher in the Institute of Industrial Science at the University of Tokyo, Japan. In 2012, he was a visiting scholar in University College London, the U.K. His research interests include GNSS positioning in challenging environments and localization for pedestrian, autonomous driving vehicle and unmanned aerial vehicle.